\documentclass[runningheads,a4paper,oribibl,envcountsame]{llncs}

  \newdimen\paravsp  \paravsp=1.3ex 

\usepackage[mathletters]{ucs} 
\usepackage{latexsym,amsmath,amssymb}
\usepackage{hyperref} 
\usepackage{algorithm} 
\usepackage[noend]{algpseudocode} 
\usepackage[capitalize]{cleveref} 
\usepackage[usenames]{color} 
\usepackage{thmtools}
\usepackage{thm-restate}

\DeclareMathOperator{\argmax}{argmax}
\DeclareMathOperator{\argmin}{argmin}

\algnewcommand\algorithmicinput{\textbf{Input:}}
\algnewcommand\Input{\item[\algorithmicinput]}
\algnewcommand\algorithmicoutput{\textbf{Output:}}
\algnewcommand\Output{\item[\algorithmicoutput]}
\algnewcommand\algorithmiceffect{\textbf{Effect:}}
\algnewcommand\Effect{\item[\algorithmiceffect]}

\newcommand{\prefix}{\sqsubset}
\newcommand{\prefixeq}{\sqsubseteq}

\def\paradot#1{\vspace{\paravsp plus 0.5\paravsp minus 0.5\paravsp}\noindent{\bf\boldmath{#1.}}} 
\def\paranodot#1{\vspace{\paravsp plus 0.5\paravsp minus 0.5\paravsp}\noindent{\bf\boldmath{#1}}} 

\def\epstr{\epsilon}            
\def\eoe{\hspace*{\fill} $\blacklozenge\quad$} 
\def\eor{\hspace*{\fill} {\LARGE\textbullet}$\quad$} 
\AtEndEnvironment{example}{\eoe\endexample}
\AtEndEnvironment{remark}{\eor\endremark}
\def\SetR{\mathbb{R}}           
\def\SetN{\mathbb{N}}           
\def\SetB{\mathbb{B}}           
\def\E{{\mathbb E}}             

\def\cA{{\cal A}}                
\def\cE{\cal{E}}                 
\def\cH{\cal{H}}                 
\def\cF{\cal{F}}                 
\def\cM{\cal{M}}                 
\def\l{\ell}                    

\iftrue\def\h{\text{\rm\ae}}         
\else\def\h{h}\fi               
\def\prenu{\nu_0}
\def\Cyl{\mathfrak{C}}
\def\Cint{\mathchoice{\sf C\!\!\!\!\!\!\int}{\sf c\!\!\!\!\int}{\sf c\!\!\!\!\int}{\sf c\!\!\!\!\int}}

\begin{document}


\title{Value Under Ignorance in \\ Universal Artificial Intelligence}

\author{Cole Wyeth\inst{1} \and
Marcus Hutter\inst{2,3}}

\institute{Cheriton School of Computer Science \\
University of Waterloo\\
200 University Ave W, Waterloo, ON N2L 3G1, Canada \\
\email{cwyeth@uwaterloo.ca}
\and
Google DeepMind \\ 
London N1C 4AG \\ 
United Kingdom
\and
School of Computing, Australian National University \\
Canberra, ACT 2601 \\ 
Australia}

\maketitle

\begin{abstract}
We generalize the AIXI reinforcement learning agent to admit a wider class of utility functions. Assigning a utility to each possible interaction history forces us to confront the ambiguity that some hypotheses in the agent's belief distribution only predict a finite prefix of the history, which is sometimes interpreted as implying a ``chance of death'' equal to a quantity called the semimeasure loss. This death interpretation suggests one way to assign utilities to such history prefixes. We argue that it is as natural to view the belief distributions as imprecise probability distributions, with the semimeasure loss as total ignorance. This motivates us to consider the consequences of computing expected utilities with Choquet integrals from imprecise probability theory, including an investigation of their computability level. We recover the standard (recursive) value function as a special case. However, our most general expected utilities under the death interpretation cannot be characterized as such Choquet integrals. 

\end{abstract}

\section{Introduction}\label{sec:Intro}

The AIXI reinforcement learning (RL) agent \cite{hutter_theory_2000} is a clean and nearly parameter-free description of general intelligence. However, because of its focus on the RL setting it does not natively model arbitrary decision-theoretic agents, but only those that maximize an external reward signal. A generalization to other utility functions, provided in this paper, is interesting for decision theory and potentially important for AI alignment. We further show how our generalization of AIXI naturally leads to imprecise probability theory, while assigning utilities to events in an extended space under a certain associated probability distribution returns us to the domain of von Neumann-Morgenstern rationality \cite{von_neumann_theory_1944}.  

\paradot{Motivation}
The AIXI policy decision-theoretically maximizes total expected returns (discounted reward sum) over its lifetime with respect to the universal distribution, which has the potential to encode essentially arbitrary tasks. Arguably, AIXI's drive to maximize expected returns would lead it to instrumentally seek power even at the expense of its creators, no matter what form of rewards they might choose to administer \cite{Cohen_Hutter_Osborne_2022}. Therefore, it is reasonable to seek a more general class of agents whose terminal goals \emph{are} parameterized \cite{beren_agi_2023}. Indeed, choosing the expected returns as optimization target was motivated by the promise of reinforcement learning, while the primary (pre-)training method for frontier AI systems is now next-token prediction \cite{hou_does_2024}, though RL still plays an important role \cite{bandyopadhyay_thinking_2025}. 
From a general decision-theoretic standpoint, we should allow as wide a class of utility functions as possible\footnote{Perhaps this is also preferable for modeling human cognition.} - and for AI alignment, we may desire a modular and user-specifiable utility function. 

Our focus is on the history-based setting of universal artificial intelligence, where agents learn to pursue their goals by exchanging actions and percepts with the environment; but without including rewards in general, we move beyond the RL paradigm. Because we are interested in \emph{universal} agents that can succeed across a vast array of environments, we cannot rely on common simplifying assumptions (such as the Markov property) and it is difficult to prove objective optimality or convergence guarantees. We cannot even rely on the usual additivity of probabilities, but are forced to work with ``defective'' \emph{semimeasures} (\cref{def:semimeasure}). In this work, we aim to develop the mathematical tools to rigorously extend AIXI to more general utility functions and investigate the properties (e.g. computability level) that carry over from the standard case.

\paradot{Main contribution}
We introduce the basics of semimeasure theory intended to model filtrations with a chance of terminating at a finite time. In the context of history-based reinforcement learning \cite{hutter_theory_2000}, we prove equivalence between the Choquet integral of the returns (with respect to the history distribution) and the recursive value function. This suggests a generalized version of AIXI which optimizes any (continuous) utility function. We prove the existence of an optimal policy under the resulting generalized value functions, and investigate their computability level, obtaining slightly better results for the Choquet integral than the ordinary expected utility. We use these results to analyze the consequences of treating semimeasure loss as a chance of death.

\section{Background}\label{sec:related_work}

\paradot{AIXI}
Our exposition of AIXI is sufficient to understand our results but does not go into detail because AIXI is the standard approach to general reinforcement learning with many good introductions available in the literature \cite{hutter_theory_2000, Hutter:04uaibook}.

\paradot{Utility functions for AIXI}
There have been several suggestions for alternative utility functions in AIXI-like models \cite{alexmennen_utility-maximizing_2012, hibbard_model_based, orseau_universal_2014}. Orseau's (square/Shannon) knowledge seeking agents (KSA) are a particularly interesting example, motivated to explore by an intrinsic desire for surprise. His general setting of universal agents $\text{A}^\rho$ hints at (but does not rigorously develop) a wide class of utility functions. Our work subsumes all of these examples. Computability aspects of (slight) variations on AIXI's value function have been investigated by Leike et al. \cite{leike_computability_2018}. We seem to be the first to rigorously formulate a general class of utility functions and show how to define optimal policies in terms of the mathematical expectation of utility in the history-based RL framework.

\paradot{Relating utility functions and rewards}
 The classical RL literature contains extensive discussions of Sutton's reward hypothesis \cite{sutton_reward_2004}``That all of what we mean by goals and purposes can be well thought of as maximization of the expected value of the cumulative sum of a received scalar signal (reward)." In particular, see \cite{bowling_settling_2023} for axioms on a preference ordering that allow it to be represented by a (discounted) reward sum. Typically, these results are based on Markov assumptions that do not apply to our setting.

\paranodot{The semimeasures}
appearing in the literature are actually only pre-semimeasures that have not been explicitly extended to a $\sigma$-algebra. Despite extensive applications of (pre-)semimeasures in AIT \cite{li_introduction_2008, hutter_theory_2000, Hutter:04uaibook, Hutter:24uaibook2}, we are not aware of any published work on the formal theory of semimeasures on Cantor space beyond the conjectures and suggestions of Hutter et al. \cite[Sec.2.8.2]{Hutter:24uaibook2}. For instance, the paragraph on ``Including Finite Sequences'' suggests the semimeasure extension approach that we follow in this work (our $P_\nu$ is their $\tilde{\nu}$), among various other conjectured approaches that we have not pursued (or not as far). See also section 2.8.4 exercise 4 for a more explicit formula. Their ``Expectations w.r.t. a semimeasure'' are (in one possible formalization) a special case of the Choquet integral which we relate to our own (extended space) integral in \cref{lemma:ext_int_to_choquet}. In the context of imprecise probability theory, there has been extensive work on ``non-additive set functions'' with much weaker assumptions than semimeasures e.g. \cite{choquet_theory_1954, gilboa_additive_1994, gilboa_canonical_1995}; in fact one of our (easier) results (\cref{lemma:ext_int_to_choquet}) can also be derived as a special case of results from \cite{gilboa_canonical_1995}. Recently, imprecise probability has been adapted for ``robust'' reinforcement learning \cite{appel_regret_2025}.

\paradot{Semimeasure loss as death}
The semimeasure loss has been interpreted as a chance of death \cite{martin2016death}. Martin et al. study various consequences, including rigorously identifying semimeasure loss with a chance of transitioning to a zero reward absorbing state, an agent's long run beliefs about its chances of death, and the suicidal behavior of agents with negative reward sets.

\section{Mathematical Preliminaries}\label{sec:math_prelim}

\paradot{Notation}
For any finite alphabet $\cA$, we denote the set of finite strings over $\cA$ as $\cA^*$. If $s \in \cA^*$, then $s_i \in \cA$ is the $i^\text{th}$ symbol of $s$, indexing from 1. Similarly $s_{i:j}$ for $i \leq j$ is the substring from indices $i$ to $j$. The empty string is written $\epstr$. The concatentation of two strings (or a string and an infinite sequence) $x$ and $y$ is denoted $xy$. We will use $x \prefix y$ to denote that $x$ is a proper prefix of $y$, that is $y = xz$ with $z \neq \epstr$, and similarly $x \prefixeq y$ means $x$ is a prefix of $y$ possibly equal to $y$. We write $\l(x)$ for the length of the string $x$ in alphabet symbols. For example, $x \prefix y$ implies $\l(x) < \l(y)$. The indicator function of the set $A$ is denoted $I_A$.

\paradot{The Cantor Space}
Because we are concerned with (finite and infinite) ``interaction histories'' between an agent and environment, which are time-indexed strings/sequences of ``action'' and ``percept'' symbols, we will need the topological properties of the Cantor space. The Cantor space is a topological space defined as the set of infinite sequences over some alphabet $\cA$ with the topology generated by the \emph{cylinder sets}, sets of sequences that begin in the same way. For each $x \in \cA^*$, let $x\cA^\infty = \{ x\omega | \omega \in \cA^\infty \}$. That is, the cylinder set is the set of all sequences beginning with $x$.\footnote{The more common (and short, but less evocative) notation is $\Gamma_x$, which we will typically avoid because it does not make the relevant alphabet explicit.} The set of cylinder sets over alphabet $\cA$ is $\Cyl_{\cA}$. With $\mathcal{P}(S)$ denoting the power set of $S$, we will need the injection $\imath : \cA^* \rightarrow \mathcal{P}(\cA^*)$ given by $\imath(x) = \{x\}$. Intuitively, we can focus on cylinder sets because we are concerned with agents that have preferences over events at finite times (necessarily; optimal policies to bring about tail events are commonly undefined, see \cref{ex:tail_event_caution}).

\begin{definition}[lower semicomputable]
A function $f : Y \rightarrow \SetR$ for $Y \subseteq \cA^* \cup \cA^\infty$ is \emph{lower semicomputable} (l.s.c.) if there is a computable function $\phi(y,k)$ monotonically increasing in its second argument with $\lim_{k\to\infty} \phi(y,k) = f(y)$. That is, $f$ can be computably approximated from below.\footnote{When $Y \cap \cA^\infty \neq \emptyset$ the input is provided to the Turing machine on an infinite tape, meaning that only a finite initial segment can be read for any fixed $k$. For our purposes in this paper, it is actually enough for lower semicomputability to hold on strings and sequences separately.}
\end{definition}

\begin{definition}[upper semicomputable]
A function $f : Y \rightarrow \SetR$ for $Y \subseteq \cA^* \cap \cA^\infty$ is \emph{upper semicomputable} (u.s.c.) if $-f$ is l.s.c.
\end{definition}

\begin{definition}[estimable]
A function $f : Y \rightarrow \SetR$ for $Y \subseteq \cA^* \cap \cA^\infty$ is estimable if it is both l.s.c. and u.s.c.
\end{definition}

\begin{definition}[pre-semimeasure]
A pre-semimeasure $\prenu$ (on $\cA^\infty$) is a function $\cA^*\to\mathbb{R}^+$ satisfying $\prenu(s) \geq \sum_{a \in \cA} \prenu(sa)$.
\end{definition}
We are mainly concerned with \emph{probability pre-semimeasures} which always take value $1$ on the empty string $\epstr$, and whether they can be extended to semimeasures. The value $\nu_0(s)$ is interpreted as the measure assigned to the cylinder set $s\cA^\infty$.

\begin{definition}[semimeasure]\label{def:semimeasure}
Let $(\Omega, \mathcal{F})$ be a measurable space. A semimeasure $\nu :\mathcal{F} \rightarrow \mathbb{R}^+$ is a $\sigma$-superadditive set function. That is, a semimeasure satisfies $\nu(\bigcup_{n=1}^\infty A_n) \geq \sum_{n=1}^\infty \nu(A_n)$ for disjoint $\{A_n\}_{n\in\SetN}$ and $\nu(A) \geq 0$. 
\end{definition}

Once we have extended our pre-semimeasures to semimeasures we will usually abbreviate the verbose $\nu(x\cA^\infty)$ as $\nu(x)$. No real confusion should arise because this interpretation is always intended for $\nu(x)$; our semimeasures are never viewed as assigning measure to finite strings. 

\begin{definition}[semimeasure loss]
    The semimeasure loss at string $x \in \cA^*$ for semimeasure $\nu$ on $\big(\cA^\infty,\sigma(\Cyl_{\cA})\big)$ is denoted
    \begin{equation*}
        L_\nu(x) ~:=~ \nu(x) - \sum_{a \in \cA} \nu(xa) \geq 0
    \end{equation*}
\end{definition}

\paradot{History-based reinforcement learning}
The pre-semimeasures that we are most interested in come from the interaction between an agent's policy and an environment. The environment may be defective (interaction may terminate at a finite time). This means the environment should give us a pre-semimeasure on percept sequences for each sequence of actions. The \emph{true} environment is denoted $\mu$, but we generally work with an abstract ``environment'' $\nu$ that can also refer to the agent's (subjective) belief distribution, usually a Bayesian mixture. Formally, the finite alphabets we will discuss are the set of actions $\cA$ available to an agent and the set of percepts $\cE$ that the environment $\nu$ might produce and send to the agent. A percept consists of an observation $o_t \in \mathcal{O}$ and (possibly) a reward $r_t \in \mathcal{R} \subset \mathbb{R}$. The action/percept at time $t$ will be denoted $a_t/e_t$, and the history of actions and observations up to time $t$ will be written $\h_{1:t} = a_1e_1...a_te_t$. When there is a reward, $e_t = o_tr_t$. For brevity, the alphabet $\cH := \cA \times\cE$. An environment is a function $\nu : \mathcal{H}^* \rightarrow [0,1]$ satisfying the chronological semimeasure condition:
\begin{equation*}
    \forall e_{1:t}, \forall a_{1:t+1}:~ \nu(\h_{1:t}) \geq \sum_{e_{t+1} \in \cE} \nu(\h_{1:t+1})
\end{equation*}
When this condition is satisfied, we write $\nu(e_{1:t} || a_{1:t}) := \nu(\h_{1:t})$ to denote the probability that the environment produces percepts $e_{1:t}$ when the agent takes actions $a_{1:t}$. 
The expression $\nu(e_{1:t} || a_{1:t})$ can be broken down into the product $\prod_{i=1}^t \nu(e_i|\h_{<i}a_i)$ where $\nu(e_i|\h_{<i}a_i) := \nu(e_{1:i} || a_{1:i}) / \nu(e_{<i} || a_{<i})$.

\section{Semimeasure Extension} \label{sec:extension} 

In order to define expected values for general utility functions, we need a rigorous notion of integration. We are particularly interested in expected utilities which can be expressed as a type of integral with respect to a semimeasure, but defining these integrals requires the pre-semimeasures ordinarily discussed in the literature to be extended to a full $\sigma$-algebra. In this section we explain how to do this by taking advantage of ordinary measure-theoretic results.

\paradot{Pre-semimeasures}
Consider a probability pre-semimeasure $\prenu$ defined on cylinder sets over alphabet $\cA$. 
The intended interpretation of $\prenu$ is that $\prenu(x)$ is the probability of a sequence starting with $x$, and $\prenu(x) \geq \sum_{a \in \cA} \prenu(xa)$ because all sequences starting with $xa$ also start with $x$. The inequality may be strict when there is a chance that the sequence terminates after the first $\l(x)$ symbols. We will take the values of $\prenu$ on cylinder sets as given, assuming that $\prenu$ is superadditive and $\prenu(\epstr) = 1$, and show that there is a unique underlying probability measure $P$ over $\Omega' = \cA^* \cup \cA^\infty$ and the $\sigma$-algebra $\mathcal{F}' = \sigma(\imath(\cA^*), \Cyl_{\cA})$ that matches this interpretation of $\prenu$. This will allow us to uniquely extend $\prenu$ to a true semimeasure $\nu$ corresponding to $P$ over $(\Omega,\mathcal{F}) := \big(\cA^\infty,\sigma(\Cyl_{\cA})\big)$. 

\begin{restatable}[Semimeasure extension]{theorem}{semimeasure}\label{thm:semimeasure_extension}
    A probability pre-semimeasure $\nu_0$ on $\cA^*$ defines a unique probability measure $P$ on $(\Omega',\mathcal{F}')$ satisfying
    \begin{align}\label{eq:P_for_nu}
    \prenu(x) ~&=~ \textstyle\sum_{y \in \cA^*} P(xy) + P(x\cA^\infty) \\
    \text{Furthermore}~~~~~
    \nu(S) ~&:=~ \textstyle\sum_{x\cA^\infty \subseteq S} P(x) + P(S) \label{eq:nu_from_P}
    \end{align}
    defines an extension of $\prenu$ to $(\Omega,\mathcal{F})$, with $P(x) = L_\nu(x)$.
\end{restatable}

\paradot{Proof idea}
This is a routine application of Carath\'eodory's extension theorem. It is well known that a probability pre-measure which is $\sigma$-additive on the cylinder sets can always be extended uniquely to $\cF$; here we weaken the requirement to superadditivity and obtain an extension which may not be unique (but is the only one matching our intended interpretation for $\nu$). 

\begin{example}[A simple defective semimeasure]\label{ex:rho}
    Consider binary alphabet $\SetB$ and $\nu^d_0(\epsilon) = 1$ but $\nu^d_0(x) = 2^{-l(x)-1}$ for $l(x) > 0$. We obtain $P(\epsilon) = 1/2$, $P(\SetB^\infty) = 1/2$. Identifying $\SetB^\infty$ with $[0,1]$ by treating sequences as binary expansions, $\nu^d(S)$ is equal to $\lambda(S)/2$ (where $\lambda$ is the Lebesgue measure) everywhere except $\nu^d([0,1]) = 1$.
\end{example}

\paradot{Termination semimeasures}
In the rest of this paper, a semimeasure obtained from a pre-semimeasure  according to \cref{eq:nu_from_P} is called a \emph{termination} semimeasure. When we wish to emphasize this connection between a specific $\nu$ and $P$ we will elaborate by replacing $P \rightarrow P_\nu$. Note that $\nu$ inherits continuity from above from $P_\nu$, but is not necessarily continuous from below.

\section{Prediction with Semimeasures}\label{sec:prediction_with_semimeasures}

Solomonoff \cite{solomonoff_formal_1964} proposed a ``universal'' prediction method which can be expressed as a Bayesian mixture of all l.s.c. semimeasures:
\begin{equation}
    \xi_U = \sum_{\nu \in \cM^\text{semi}_\text{lsc}} w_\nu \nu
\end{equation}
In order to approximate this method (to lower semi-compute $\xi_U$) we must at least be able to list the predictors in $\cM^\text{semi}_\text{lsc}$. We can do this because each predictor is given by a probabilistic program. Unfortunately, there is no algorithm to determine which programs compute probability measures; we are unable to ``filter out'' semimeasures with nonzero semimeasure loss. In this sense we are forced to consider \emph{all} l.s.c. semimeasures as predictors. A similar problem occurs for AIXI, which uses an interactive version of $\xi_U$ for its belief distribution (explained in \cref{def:aixi}); for the moment we will focus on $\xi_U$ for simplicity. 

\paradot{Interpreting semimeasure predictors}
We can choose a more or less literal interpretation of these (strict) semimeasure predictors. If we view the semimeasure loss as a chance of sequence termination, then the semimeasure predictor $\nu$ can be taken literally as a generative hypothesis for the sequence. The true environment can be expressed by some $P_\mu$ of form \cref{eq:nu_from_P}, which may place nonzero measure on finite sequences. Alternatively, we may view the semimeasure loss as a defect of our predictors. This view treats the Bayesian mixture as a pragmatically effective method of aggregating the predictions of each $\nu \in \cM^\text{semi}_\text{lsc}$, without assuming that any $\nu$ is necessarily \emph{true} (see \cite{sterkenburg_universal_2018} for an engaging discussion of this distinction). This implies that the semimeasure loss $L_\nu(x)$ arises from $\nu$ failing to produce further predictions after $\nu$, which leads the predictive probabilities of $\xi_U$ to be non-additive. $\xi_U$ could perform better by redistributing the lost measure, but there does not seem to be a canonical (let alone l.s.c.) method for achieving this. The most common proposal is the Solomonoff normalization, which (roughly speaking) ignores any $\nu$ that does not yield predictions and uses all predictions available at each step, reweighted so that the conditionals add up to 1. 

\paradot{Semimeasures as credal sets}
Imprecise probability is sometimes advocated as a generalization of Bayesian decision theory to deal with model misspecification or unrealizability, dispensing with the assumption that the hypothesis class contains the truth \cite{walley_statistical_1991}.
Instead, advocates view knowledge as ``incomplete information'' that is insufficient to fully specify a probability distribution, assuming total \emph{ignorance} between possibilities which cannot be reduced to meaningful probability assignments. A (probability) semimeasure $\nu$ is translated to a whole family of ``possible'' probability distributions (or a \emph{credal set}) called $\text{Core}(\nu)$, which contains all probability measures $p$ with $p(A) \geq \nu(A)$. Informally, the core of a termination semimeasure includes probability measures that allocate the semimeasure loss $L_\nu(x) = P_\nu(x)$ arbitrarily to support $x\cA^\infty$. This is the set of possible predictors we can obtain by ignoring those $\nu$ that stop producing predictions.

\paradot{Implications for universal agents}
Generalizing the utility function of AIXI forces us to choose an interpretation. If we take semimeasures loss as termination (or ``death'') we must assign utilities to finite histories. Otherwise, we need to either renormalize or choose a decision rule appropriate for non-singleton credal sets. Formally, this depends on how we define semimeasure integrals and the resulting value functions. We are not aware of strong arguments that the semimeasure loss tends to coincide with a chance of death, and we will see that the credal set interpretation can have better computability properties.

\section{Value Functions and Integration}\label{sec:expectations_and_integration}

Now we are prepared to define integration with respect to a semimeasure. We characterize semimeasure integrals in terms of an integral against the associate measure and investigate their connection to the classical value functions of history-based reinforcement learning. 

\paradot{Recursive value function}
Our original inspiration to define semimeasure integration is to express (and generalize) the ``recursive'' value function \cite[Def.4.10]{leike_nonparametric_2016} from history-based reinforcement learning:
\begin{equation}\label{eq:rec_V}
    V^\pi_\nu ~=~ \sum_{t = 1}^\infty \sum_{\h_{1:t}} \gamma_t r_t \nu^\pi(\h_{1:t})
\end{equation}
for $\gamma_t \geq 0$, $\sum_{t=1}^\infty \gamma_t < \infty$.\footnote{Often, $V^\pi_\nu$ is normalized by a factor of $(\sum_{t=1}^\infty \gamma_t)^{-1}$.} $V^{\pi}_{\nu}$ can be seen as an expected value of $\sum_{t=1}^\infty \gamma_t r_t $, which could in principle be replaced by another ``utility function'' $u$ to yield an expected utility $V^{\pi}_{\nu,u} = \mathbb{E}_{\nu^\pi}[u]$. When $\nu^\pi$ is a (defective) semimeasure, it is not immediately clear how to make the right-hand side rigorous, which we will take up below. 

\begin{example}\label{ex:perilous_env}
    Consider the ``perilous'' environment $\nu^p$ with action set $\cA = \{1,2\}$ that always yields empty observation and reward $r_t = a_t$, and terminates with chance $1/2$ when action 2 is chosen. Let $\gamma_t = 2^{-t}$. Then the policy $\pi_2$ that deterministically chooses action 2 has value
    \begin{equation*}
        V^{\pi_2}_{\nu^p} ~=~ 2 \cdot \frac{1}{4} \cdot \frac{1}{1 - 1/4} = \frac{2}{3}
    \end{equation*}
    by the standard formula for a geometric series. The policy $\pi_1$ that selects action 1 has value
    \begin{equation*}
        V^{\pi_1}_{\nu^p} ~=~ 1 \cdot \frac{1}{2} \cdot \frac{1}{1 - 1/2} = 1
    \end{equation*} \\[-5ex]
\end{example}

We will develop semimeasure integration so as to be able to write $V_{\nu}^{\pi}$ as a $\nu^\pi$-expectation of $\sum_{t=1}^\infty \gamma_t r_t$, allowing us to naturally replace the returns with alternative choices of utility function.  

Given a probability measure $\nu$ with cumulative distribution function $F$, 
ordinary probability allows us to calculate the expectation of function $f:\Omega\to\SetR^+$  as
    \begin{equation*}
        \begin{split}
            \E_\nu f ~&=~ \int_0^\infty (1 - F(b)) db 
            = \int_0^\infty \nu(f > b) db \\
        \end{split}
    \end{equation*}
The term $\nu(f > b)$ can be replaced by $\nu(f \geq b)$ without changing the integral because $f$ can only take countably many values with positive probability. This formula can naturally be extended to the semimeasure case, defining the Choquet integral \cite{choquet_theory_1954}:

\begin{definition}[\boldmath Choquet integral $\nu$]\label{def:choquet}
    The Choquet integral of a measurable function $f : \Omega \rightarrow \SetR$ with respect to a semimeasure $\nu$ is given by
    \begin{equation}
        \Cint fd\nu ~:=~ \int_0^\infty \nu(f \geq b) db + \int_{-\infty}^0 [\nu(f \geq b) - \nu(\Omega)]db 
    \end{equation}
    It is well-defined and finite for bounded $f$ and well-defined but possibly infinite for non-negative $f$.
\end{definition}

For probability measures,
\begin{equation}\label{eq:graph_partition}
    \nu(\Omega) = \nu(f \geq b) + \nu(f < b)
\end{equation}
and the Choquet integral is equivalent to the $\nu$-expectation even when $f$ takes negative values. Since we are focused on probability semimeasures we may assume that $\nu(\Omega) = 1$, but \cref{eq:graph_partition} may not hold. 

When $\nu = \mu^\pi$ and $f(\h) = \sum_{t=1}^\infty \gamma_t r_t$, we recover the recursive value function, as stated in the following theorem:

\begin{theorem}[\boldmath $V^\pi_\mu = \Cint (\Sigma_t \gamma_t r_t) d\mu^\pi$]\label{thm:E_to_V}
    Let $\Omega := {\cH}^\infty = (\cA \times \cE)^\infty$, where $\mathcal{E} \ni e_t = o_tr_t$ with $r_t \in \mathcal{R} \subset \SetR^+$, and $\mathcal{R}\ni 0$ is a finite set of rewards. Then $V^\pi_\nu = \Cint (\Sigma_t \gamma_t r_t) d\nu^\pi$.
\end{theorem}

The proof relies on the following lemma, which is interesting in its own right.

\begin{lemma}[\boldmath $\int \underline{f}dP_\nu = \Cint f d\nu$]\label{lemma:ext_int_to_choquet}
    Consider the (semi)probability space $(\Omega, \mathcal{F}, \nu)$ where $\Omega = \mathcal{A}^\infty$, $\mathcal{F} = \sigma(\Cyl_{\cA})$, and $\nu$ is a termination semimeasure. Let $f : \Omega \rightarrow \SetR^+$ be an $\mathcal{F}$-measurable function. Then
    \begin{equation*}
        \Cint f d\nu ~=~ \int \underline{f}dP_\nu
    \end{equation*}
    where $\underline{f} : \Omega' \rightarrow \SetR^+$ is defined by $\underline{f}(\omega) = f(\omega)$ and $\underline{f}(x) = \inf_{\omega \in x\cA^\infty} f(\omega)$.
\end{lemma}

This easy result is a special case of \cite[Thm.4.3]{gilboa_additive_1994} for finite sample spaces. See \cite[Thm.E]{gilboa_canonical_1995} for the infinite case.

\paradot{Choquet integral as minimum over credal set}
By \cite[Thm.2.1\&2.2]{gilboa_canonical_1995}, for bounded, measurable $f$ and convex semimeasure $\nu$ (including our termination semimeasures),
\begin{equation*}
    \Cint f d\nu = \min_{p \in \text{Core}(\nu)}\int f dp
\end{equation*}
In the context of expected utility, this means the Choquet integral is pessimistic (and the corresponding decision rule is max-min). This explains why $\Cint (\Sigma_t r_t) d\nu^\pi$ almost coincidentally equals $V_\nu^\pi$: the worst possible history in $\h_{1:t}\cH^\infty$ has value $\min_{r_{t+1:\infty}} \sum_{i=1}^\infty \gamma_i r_i = \sum_{i=1}^t \gamma_i r_i$, so $\argmin_{p\in\text{Core}} \int (\Sigma_i r_i)dp$ concentrates the measure $L_{\nu^\pi}(\h_{1:t})$ on histories with $r_{t+1:\infty} = 0$. As noted by \cite{martin2016death}, transitioning to an absorbing death state with 0 reward is the same as terminating the interaction and awarding the agent the corresponding \emph{partial} sum of discounted reward, which is the intended semantics of the recursive value function. Therefore, the Choquet integral is equivalent for essentially the same reason (but without an explicit death state).\ 

\paradot{Note on the perilous environment}
In \cref{ex:perilous_env}, the reward set does not contain 0, so the Choquet integral $\int (\Sigma_t \gamma_t r_t) d{(\nu^p)}^{\pi_2} > V^{\pi_2}_{\nu^p}$.  

\section{AIXI with General Utility Function}\label{sec:aixi}

With the mathematical equipment defined above, we can generalize the AIXI model to pursue many utility functions beyond the given reward sum \cite{alexmennen_utility-maximizing_2012,hibbard_model_based} and we may do away with rewards completely, simplifying environments to only produce observations. We show under which conditions this generalization is well-defined and make an introductory investigation of its computability properties. In the greatest generality, we can assign utilities to both finite (terminating) and infinite histories. However, we must require a type of continuity from our utility function if we want an optimal policy to exist:

\begin{definition}[Continuity]\label{def:continuity}
    The function $f : \cA^\infty \rightarrow \mathbb{R}$ is continuous (with respect to the Cantor space topology on $\cA^\infty$ and the standard topology on $\mathbb{R}$) if for any $\omega \in \cA^\infty$, 
    \begin{equation*}
        f(\omega) ~=~ \lim_{n\rightarrow\infty}\ \inf_{\omega' \sqsupseteq \omega_{1:n}} f(\omega') = \lim_{n\rightarrow\infty} \ \sup_{\omega' \sqsupseteq \omega_{1:n}} f(\omega') 
    \end{equation*}
    and similarly we say that $f : \cA^* \cup \cA^\infty \rightarrow \mathbb{R}$ is continuous if
    \begin{equation*}
        \begin{split}
           f(\omega) ~&=~ \lim_{n\rightarrow\infty} \min\{\inf_{\omega' \sqsupseteq \omega_{1:n}} f(\omega'), \min_{x \sqsupseteq \omega_{1:n}} f(x)\} \\
           ~&=~ \lim_{n\rightarrow\infty} \max\{\sup_{\omega' \sqsupseteq \omega_{1:n}} f(\omega'),\max_{x \sqsupseteq \omega_{1:n}} f(x)\}  
        \end{split}
    \end{equation*}
    The later continuity notion is induced by the topology on $\Omega' := \cA^* \cup \cA^\infty$ generated by the open sets $A_x=x\Omega'$.
\end{definition}

\begin{definition}[Utility-based AIXI]\label{def:aixi}
     Consider a continuous utility function $u : \cH^* \cup \cH^\infty \rightarrow \mathbb{R}^+$. Recalling that $\nu^\pi$ is the semimeasure induced by the interaction of chronological semimeasures $\nu$ and $\pi$, let $P_{\nu^\pi}$ be the corresponding measure, so that for all $S \in \mathcal{F} = \sigma(\Cyl_{\cH})$,
     \[
        \nu^\pi(S) ~= \sum_{\h_{\leq t}{\cH^\infty} \subseteq S} P_{\nu^\pi}(\h_{\leq t}) + P_{\nu^\pi}(S)
     \]
     We define $V^\pi_{\nu,u} = \int u\,dP_{\nu^\pi}$, a standard Lebesgue integral. Let $\xi^\text{AI}$ be a universal mixture of the set $\cM^\text{ccs}_\text{lsc}$ of l.s.c.\ chronological semimeasures \cite{Hutter:04uaibook},
     \begin{equation}
         \xi^\text{AI} ~:= \sum_{\nu\in\cM^\text{ccs}_\text{lsc}} w_\nu \nu
     \end{equation}
     where conventionally $w_\nu := 2^{-K(\nu)}$, $K$ denoting the prefix Kolmogorov complexity, an explicit ``Occam's razor''-style complexity penalty.
     Then  
     \begin{equation*}
         \pi^\text{AIXI} ~:=~ \pi^*_{\xi^\text{AI}} ~:=~ \argmax_\pi V^\pi_{\xi^\text{AI},u}
     \end{equation*}
\end{definition}

\begin{example}[Caution]\label{ex:tail_event_caution}
    There are utility functions that have no optimal policy. For instance, consider the environment with binary action set $\cA = \SetB$ which only yields nonzero reward on the first time step $t$ that action 1 is taken, but in that case the reward is $1 - 1/t$. Also assume that there is no discounting ($\gamma_t = 1$, violating the assumptions of the recursive value function). This is a well-defined utility function, but it is not continuous, and it has no optimal policy: though it is certainly wise to take action 1 eventually, it is always better to do it one time step later.  
\end{example}
In light of the previous example, we must show that \cref{def:aixi} makes sense. Fortunately, existence of $\pi^\text{AIXI}$ follows from compactness of Cantor space and continuity of $u$, by an argument similar to \cite[Thm.8]{lattimore_general_2014}. In the case of value functions representable as Choquet integrals, the (hyper)computability properties of $V^\pi_\nu$ carry over to our general setting as well:

\begin{theorem}[\boldmath $V^\pi_{\nu,\underline{u}}$ lower semicomputability ]\label{thm:main}
    Let $u : \cH^\infty \rightarrow \SetR^+$ be l.s.c. and continuous.\footnote{Lower semicomputability only implies lower semicontinuity in general.} Then $\underline{u}$ is l.s.c. and continuous. If $\nu^\pi$ is also l.s.c. then $V^\pi_{\nu,\underline{u}} = \Cint u\,d\nu^\pi$ is l.s.c., and if $\nu$ is l.s.c. then $V^*_{\nu,\underline{u}} := V^{\pi^*_\nu}_{\nu,\underline{u}} = \max_\pi \Cint u\,d\nu^\pi$ is l.s.c.
\end{theorem}

\paradot{Proof idea} 
Kőnig's lemma allows $\underline{u}$ to inherit lower semicomputability from $u$. Then $\underline{u}$ can be approximated from below on the cylinder sets $\h_{1:n}\cH^\infty$ by l.s.c. simple functions $\underline{u_n}$ (taking $u_n(\h) = \underline{u}(\h_{1:n})$), and $V^\pi_{\nu,\underline{u_n}} \rightarrow V^\pi_{\nu,\underline{u}}$ by the monotone convergence theorem. By \cref{thm:E_to_V}, $\forall n,\ V^\pi_{\nu,\underline{u_n}} = \Cint \underline{u_n}d\nu^\pi$ which is directly l.s.c. It is easy to see that there is always a deterministic optimal policy $\pi^*_\nu$ (indeed, even when defective semimeasure policies are allowed, they are never preferred to proper measures against $\underline{u}$). Therefore the optimal value function can be approximated by dovetailing the calculations of $V^\pi_{\nu,\underline{u}}$ for each (deterministic) $\pi$ and tracking the running maximum. 

\paradot{Note on Choquet integral form}
It is essential for the proof of lower semicomputability that $V^\pi_{\nu,u}$ can be expressed in terms of a Choquet integral ($u = \underline{u_0}$ for some $u_0$), and indeed the value function may not be l.s.c. otherwise. For example, consider the standard value function under the death interpretation of semimeasure loss, which sums the discounted rewards received over the (possibly terminating) history: $u_r(\h) := \sum_{t=1}^{l(\h)} \gamma_t r_t$ for $\h \in \cH^* \cup \cH^\infty$. Assuming $\sum_{t=1}^\infty \gamma_t < \infty$ (usually scaled $\leq 1$), $u_r$ is continuous and l.s.c. (even estimable). However, when $\inf\mathcal{R}<0$ and $\gamma_t>0\,\forall t$, then $u_r({\h}_{\leq t}) \neq \inf_{\omega \in \h_{\leq t} \cH^\infty} u_r(\omega)$, and the standard value function $V^\pi_{\nu,u_r}$ is not l.s.c. in general. For example, inverting a positive set of rewards including 0 produces a u.s.c. value function which is typically not l.s.c. When $0 \in \mathcal{R} \subset \SetR^+$, $V^\pi_{\nu,u_r} = V^\pi_{\nu, \underline{\Sigma_t \gamma_t r_t}}$ by \cref{thm:E_to_V}, implying lower semicomputability by \cref{thm:main}. 

\paradot{Implications}
While this nice relationship seems to indicate that value functions are (most) naturally represented as Choquet integrals, there are a couple of important caveats. Only the weaker notion of limit computability for $V^\pi_\nu$ is used by \cite{leike_computability_2018} to obtain limit computable $\varepsilon$-approximations to AIXI, so lower semicomputability does not directly improve this result. However, l.s.c. of $V^\pi_{\nu,u_r}$ (when it holds) can be used to show convergence of AIXI$tl$ \cite[Thm.13.2.4]{Hutter:24uaibook2} depending on the strength of the proof system \cite[Sec.5.2.3]{leike_nonparametric_2016}. This situation applies similarly to $V^\pi_{\nu,\underline{u}}$. While probabilistic conditioning (that is, Bayesian updating) often breaks lower semicomputability of $\nu^\pi$, including for $(\xi^\text{AI})^\pi$ when $\pi$ is computable, any policy maximizing $V^\pi_{\nu,\underline{u}}$ remains optimal at all times except with $\nu^\pi$ probability 0. Also, after observing $\h_{\leq t}$ it is sufficient to maximize the ``renormalized value function'' $\Cint_{\h_{\leq t}\cH^\infty} u d\nu^\pi := \Cint uI_{\h_{\leq t}\cH^\infty} d\nu^\pi$ fixing $\pi(\h_i|\h_{< i}) = 1$ for $i \leq t$.

\section{Discussion}\label{sec:disc}

We have studied the structure of value functions in universal artificial intelligence, and generalized them substantially by replacing the returns  with any continuous utility function. Interpreting the epistemology and motivations of the resulting agents requires us to reevaluate the semantics of sequence termination and the resulting semimeasure loss. The ``mainstream'' view is to take termination very literally and equate it with the agent's death. We have proposed treating semimeasures as specifying credal sets instead, evaluating expected utilities with the Choquet integral, and considered some of the consequences. Surprisingly, this improves the hypercomputability level of the value function and can recover the original recursive value function as a special case. Such Choquet integral value functions can also be viewed as expected utilities of a corresponding extended utility function.

Our analysis has been exploratory and not very deep, leaving many open questions. While there are problems for the death interpretation, our alternative view of semimeasures as imprecise probabilites does not inherently justify pessimism in the face of ignorance, as in the Choquet integral. We might instead search for a philosophically justified normalization, following Solomonoff. In future work, we intend to investigate an even larger class of utility functions with hypercomputability levels higher in the arithmetic hierarchy. 

\paradot{Acknowledgments}
This work was supported in part by a grant from the Long-Term Future Fund (EA Funds - Cole Wyeth - 9/26/2023). We would like to thank Vanessa Kosoy for helpful discussions on imprecise probability in the context of sequential decision theory and Tor Lattimore for generously providing feedback. Cole Wyeth would also like to thank his supervisor Ming Li for freedom and support to pursue this project.

\newpage

\bibliographystyle{alpha}

\newcommand{\etalchar}[1]{$^{#1}$}


\end{document}